\documentclass{article}

\usepackage{arxiv}

\usepackage[utf8]{inputenc} 
\usepackage[T1]{fontenc}  
\usepackage{hyperref}    
\usepackage{url}      
\usepackage{booktabs}    
\usepackage{amsfonts}    
\usepackage{nicefrac}    
\usepackage{microtype}   
\usepackage{lipsum}
\usepackage{pgfplots} 
\pgfplotsset{width=12cm,compat=1.9} 

\usepackage{float} 

\title{The Czech Court Decisions Corpus (CzCDC): Availability as the First Step}

\author{
 Tereza Novotná \\
 Institute of Law and Technology, Faculty of Law\\
 Masaryk University\\
 Brno, Czech Republic \\
 \texttt{tereza.novotna@mail.muni.cz} \\
  \And
Jakub Harašta \\
 Institute of Law and Technology, Faculty of Law\\
 Masaryk University\\
 Brno, Czech Republic \\
 \texttt{jakub.harasta@law.muni.cz} \\
}

\begin{document}
\maketitle

\begin{abstract}
In this paper, we describe the Czech Court Decision Corpus (CzCDC). CzCDC is a dataset of 237,723 decisions published by the Czech apex (or top-tier) courts, namely the Supreme Court, the Supreme Administrative Court and the Constitutional Court. All the decisions were published between 1\textsuperscript{st} January 1993 and 30\textsuperscript{th} September 2018.

Court decisions are available on the webpages of the respective courts or via commercial databases of legal information. This often leads researchers interested in these decisions to reach either to respective court or to commercial provider. This leads to delays and additional costs. These are further exacerbated by a lack of inter-court standard in the terms of the data format in which courts provide their decisions. Additionally, courts' databases often lack proper documentation.

Our goal is to make the dataset of court decisions freely available online in consistent (plain) format to lower the cost associated with obtaining data for future research. We believe that simplified access to court decisions through the CzCDC could benefit other researchers.

In this paper, we describe the processing of decisions before their inclusion into CzCDC and basic statistics of the dataset. This dataset contains plain texts of court decisions and these texts are not annotated for any grammatical or syntactical features.
\end{abstract}


\section{Introduction}
The Czech legal system belongs to the family of continental legal systems. As such, it relies primarily on codified legal rules. The role of judicial decisions is limited compared to the United States or the United Kingdom. However, there is a general consent in the Czech legal community that judicial decisions of apex courts are argumentatively binding. This stems from the principle of legal certainty, under which a person has a right to have her cases decided similarly to previous similar cases. Because of this reasoning, court decisions are widely studied and are being used for support of legal argumentation by judges, attorneys and students alike. The outcome of decision-making of the apex courts is used by researchers and academics as a source of legal knowledge. It can be used for qualitative or quantitative analysis, e.g. to study compliance of apex courts with case law of the European Court of the Human Rights\cite{Smekal} or for comparative examination of judicial dissent\cite{Bricker}.

In spite of the high value of court decisions published by the Czech apex courts – meaning the Supreme Court, the Supreme Administrative Court and the Constitutional Court – access to these data in bulk is quite limited. To our knowledge, it is not currently possible to download bulk court data. All three courts run individual databases which allow searching by parameters (full text search, metadata search), but these databases are suitable only for consultation of individual decisions, and not for bulk download of search results. Similarly, proprietary legal information systems do not provide us with tools to access and analyse bulk data. Individual consultation of decisions still remains the \textit{modus operandi} for the majority of lawyers. In 2016, the Czech Ombudsman concluded that the government fails to provide its citizens with legal information in suitable quality and quantity.\footnote{Expert study of the Czech Ombudsman on accessibility of Czech court decisions available in Czech at \url{http://eso.ochrance.cz/Nalezene/Edit/4496}} 

Additionally, a lack of data is also problematic in the fields of machine learning, natural language processing or computational linguistics as it slows down research or experimental activities possibly leading to useful applications. There are several well established NLP tasks for legal domain, such as reference recognition, text summarisation or argument extraction. In order to make progress in automated processing of legal documents, we need manually annotated corpora and evidence that these corpora are accurate \cite{Walker}.

We aim to address this issue by publishing the Czech Court Decisions Corpus (CzCDC) in the LINDAT/CLARIN repository at \url{http://hdl.handle.net/11372/LRT-3052}. In this paper, we describe the dataset of 237,723 court decisions, its development and its basic properties. CzCDC is organised into three sub-corpora per individual courts. In Section 2 of this paper we explain the related work which we drew inspiration from. In Section 3 we describe pre-processing of data for inclusion into CzCDC. Section 4 contains basic statistics, while Section 5 concludes the paper and outlines possible future work to enhance the quality of data contained in the CzCDC.

\section{Related Work}
\label{related}
Access to selected and curated legal texts is useful in adjudication, education and legal research of legislation\cite{Vogel}. Unfortunately, access is often hindered by pre-designed user interface \cite[p. 1351]{Vogel} or restrictive data mining policies on court websites. "Freeing" the court decisions in this regard may require significant effort - this part overviews these efforts.

There is already a significant amount of datasets focused on legal domain. Non-exhaustive list includes:
\begin{enumerate}
  \item The HOLJ corpus \cite{Grover},
  \item German Juristiches Referenzkorpus \cite{Hamann}
  \item DS21 Corpus and Swiss Legislation Corpus (SLC) \cite{Hofler}
  \item The British Law Report Corpus (BLRC) \cite{Perez}
  \item Corpus of Historical English Law Reports (CHELAR) \cite{Ro-Pue}
  \item JRC-Acquis \cite{Steinberger}
  \item The United Nations Parallel Corpus \cite{Ziemski}
  \item US Case Law Access Project \cite{CaseLaw}
\end{enumerate}

Corpora focusing on the legal domain in the Czech language are understandably scarcer. JRC-Acquis \cite{Steinberger} focuses on legal documents in official languages of the European Union. As such, it contains documents in Czech. Some of the researchers engaged in automated processing of legal texts made their training datasets, used for machine learning purposes, public at LINDAT/CLARIN repository\footnote{\url{https://lindat.mff.cuni.cz/en}}. This includes Kriz et al. \cite{Kriz} and Harašta et al. \cite{Harasta} for reference recognition tasks, and Hoang and Bojar \cite{Hoang} for machine translation. The other available corpus is CzEng, Czech-English Parallel Corpus \cite{Bojar}, also used in machine translation.

Some of the abovementioned corpora serve specific purposes. Parallel corpora are used for translation \cite{Bojar,Hoang,Steinberger,Ziemski}, annotated corpora serve as training datasets for specific particular tasks, such as reference recognition \cite{Harasta,Kriz} or document summarisation \cite{Grover}. 
On the other hand, some corpora and datasets have no specific purpose beside making the data available in bulk. Their value lies either in their completeness or representativeness. These cases include, among others, datasets of historical legal documents - Swiss DS21 contains document from the year 754 and overall spans more than 1000 years \cite{Hofler}. Some datasets contain both historical and contemporary documents, such as \cite{Ro-Pue} spanning from 1535 to 1999 or \cite{CaseLaw} spanning from 1658 to 2018. Some corpora focus predominantly on contemporary legal documents, such as German Juristiches Referenzkorpus \cite{Hamann}, The British Law Report Corpus \cite{Perez}, and Swiss Legislation Corpus \cite{Hofler}.

\section{Method}
\label{method}
In this section, we describe the pre-processing of court decisions included in the CzCDC. Because the three Czech apex courts do not share a common standard for making their decisions available for searching and individual consultation, all of our steps aimed to achieving the one goal: plain text version of court decisions enriched by basic descriptive metadata.

\subsection{Data Collection}
The three Czech apex courts make their decisions available online. Decision of the Constitutional Court are available at \url{https://nalus.usoud.cz}, the Supreme Administrative Court at \url{http://nssoud.cz}, and the Supreme Court at \url{http://nsoud.cz}. Based on full text or metadata search, it is possible to consult the decisions individually. It is not possible to access all the search results and download them in bulk, or to download the whole database in bulk. That is significantly different from situation for example in the Netherlands\footnote{\url{https://www.rechtspraak.nl/}} or in the European Union \footnote{\url{https://eur-lex.europa.eu}}.

To remedy these limitations, we have filed the freedom of information requests under the Czech law. In response to these requests, the Constitutional Court provided all the decisions as a set of .RTF documents, and the Supreme Administrative Court provided all the decisions as a set of machine readable .PDF documents. Unfortunately, the Supreme Court was not able to provide the dataset free of charge. Therefore, we mined the Supreme Court database, and downloaded all their decisions in batches of several hundred decisions at once. As we know from the Expert Study of the Czech Ombudsman, the online database of Supreme Court does not contain all decisions. Specific types of decisions, such as decision on recognition of foreign judgments are not published online.\footnote{Expert study of the Czech Ombudsman available in Czech at \url{http://eso.ochrance.cz/Nalezene/Edit/4496}} Additionally, not all decisions (especially from the 1990s) appear in the public database of the Supreme Court. Understandably, these limitations apply to our corpus as well. Decisions of the Supreme Court were obtained in plain text format.

All the decisions we obtained were already anonymised and identified by their docket numbers. Docket number is a unique number consisting of identification of court or senate of the court handling the case, number assigned based on the order by which the cases arrive to the court, and by the year of delivery of the case to the court.

CzCDC currently contains documents published no later than 30\textsuperscript{th} September 2018.

\subsection{Data Unification}
Because court decisions were obtained from different sources and in different court-dependent standards, it was necessary to process them to achieve the unified standard. Our decision, motivated by the largest possible availability, was to make the dataset available in plain text. We used Apache Tika to obtain plain text of the Constitutional and Supreme Administrative Court decisions. Then, we extracted docket numbers and dates of decisions of individual cases from the resulting plain texts.

\subsection{Corpus Description}

We believe the corpus might be beneficial for both lawyers and non-lawyers (e.g. computer scientists, linguists) and for different purposes. Therefore, we selected the most basic metadata categories.

Every court decisions included in the CzCDC is described by the following metadata contained in the accompanying .CSV file:
\begin{enumerate}
  \item docket number of court decision
  \item name of the corresponding .TXT file containing the full text of the court decision
  \item date of decision
  \item identification of court publishing the decision
\end{enumerate}

Identification of court that published the decision divides the corpus into three different sub-corpora. Every decision is marked as belonging to sub-corpus by an abbreviation of the respective court: 'ConCo' for the Constitutional Court, 'SupCo' for the Supreme Court, and 'SupAdmCo' for the Supreme Administrative Court. Date when the decision was issued is provided in ISO 8601 format (YYYY-MM-DD). Docket numbers and file names of individual decisions remain in their original format.

\section{Data Statistics}
\label{stats}
All decisions included in the CzCDC were published from 1\textsuperscript{st} January 1993 to the 30\textsuperscript{th} September 2018. The corpus contains 237,723 decisions of Constitutional, Supreme and Supreme Administrative Court. The corpus contains total of 460,524,867 words. The statistics of decisions published in individual years for individual courts are shown in Annex 1. Charts allowing easy comparison of the number of documents per court per year are in Annex 2.

\subsection{Sub-corpus of the Constitutional Court}
The Constitutional Court rules on individual violations of human rights by courts and other public bodies. It is also concerned with derogation of unconstitutional acts and local bylaws. It is not part of traditional appellate structures of Czech courts, but serves as the guardian of the Constitution.

The Sub-corpus of Constitutional Court provides digitized and anonymised decisions from 1\textsuperscript{st} January 1993 until 30\textsuperscript{th} September 2018. It consists of 73,086 judicial decisions (30,75\% of the total amount of decisions). These decisions contain total of 98,623,753 words (21,42\% of the total amount of words). Comparison with commercial databases suggests our dataset contains approx. 99,5\% of all decisions of the Constitutional Court.

\subsection{Sub-corpus of the Supreme Court}
The Supreme Court is the final appellate instance for civil and criminal cases. It deals with appeals against decisions of the lower courts. As a part of the Czech court system, it also unifies the decision-making of the lower courts in civil and criminal matters.

Judicial decisions of the Supreme Court are dated from 1\textsuperscript{st} January 1993 until 30\textsuperscript{th} September 2018. The total number of decisions is 111,977 (47,1\%) and this sub-corpus contains total of 224,061,129 words (48,65\%). Comparison with commercial databases suggests our dataset contains more than approx. 91\% of all decisions of the Supreme Court.

\subsection{Sub-corpus of the Supreme Administrative Court}
The Supreme Administrative Court is the final instance for administrative matters, such as cases brought forth by asylum seekers, data protection matters, electoral complaints etc. As part of the Czech court system, it also fulfils the role of unifying the decision-making of the lower courts in administrative matters.

The Supreme Administrative Court was founded only in 2003. Therefore, court decisions are dated from 1\textsuperscript{st} January 2003 until 30\textsuperscript{th} September 2018. The Sub-corpus of the Supreme Administrative Court consists of 52,660 decisions (22,15\%). These decisions contain the of 137,839,985 words (29,93\%). Comparison with commercial databases suggests our dataset contains more than 99,9\% of all decisions of the Supreme Administrative Court.

\section{Conclusion and Future Work}
\label{conclusion}

In this paper, we described the dataset containing the decision of the three Czech apex courts. CzCDC now contains 237,723 court decisions divided into three sub-corpora accompanied by basic metadata. We are providing the data in plain text to ensure maximal accessibility. Additionally, we chose only fundamental metadata that were easy to extract from court decisions without reliability issues. Dataset is publicly available at \url{http://hdl.handle.net/11372/LRT-3052}.

This paper frames what we understand as the beginning of a longer commitment. The first step, publication of plain dataset that accompanies this paper, supplies the public with access to otherwise problematically accessible court decisions.

As part of the future development of the dataset, we suggest to include more metadata providing more information within the attached database. Suitable metadata are references to other decisions or grammatical features to establish this database as a source for legal linguistics.

We would like to finish this paper as a plea to other research teams. CzCDC in its current version makes available 237,723 court decisions, divided into three sub-corpora. Contribution from other research teams in terms of further processing either the whole corpus or one or more of the sub-corpora would lead to enrichment of CzCDC. Also, it might eventually lead to CzCDC containing the plethora of other metadata or further sub-corpora suitable for other tasks. Ultimately, it may lift the fog of inaccessibility and unavailability that currently covers the legal domain in the Czech Republic.

\section*{Acknowledgment}
T.N. and J.H. gratefully acknowledge the support from the Czech Science Foundation under grant no. GA17-20645S. T.N. also acknowledges the institutional support of Masaryk University.

\bibliographystyle{unsrt} 


\begin{thebibliography}{1}

\bibitem{Bojar}
Ondřej Bojar, Ondřej Dušek, Tom Kocmi, Jindřich Libovický, Michal Novák, Martin Popel, Roman Sudarikov, and Dušan Variš. CzEng 1.6: Enlarged Czech-English Parallel Corpus with Processing Tools Dockeres. \textit{Proceedings of TSD 2016}, pp. 231--238.

\bibitem{Bricker}
Benjamin Bricker. Breaking the Principle of Secrecy: An Examination of Judicial Dissent in the European Constitutional Courts. \textit{Law \& Policy}, 2017, vol. 39, no. 2, pp. 170--191.

\bibitem{CaseLaw}
Caselaw Access Project. 2018, \url{https://case.law/api/}.

\bibitem{Grover}
Claire Grover, Ben Hachey, Ian Hughson. The HOLJ Corpus: supporting summarisation of legal texts. \textit{Proceedings of the 5th International Workshop on Linguistically Interpreted Corpora}, 2004, pp. 47--53.

\bibitem{Hamann}
Hanjo Hamann, Friedemann Vogel and Isabelle Gauer. Computer Assisted Legal Linguistics (CAL\textsuperscript{2}). \textit{Proceedings of JURIX 2016}, pp. 195--198.

\bibitem{Harasta}
Jakub Harašta, Jaromír Šavelka, František Kasl, Adéla Kotková, Pavel Loutocký, Jakub Míšek, Daniela Procházková, Helena Pullmannová, Petr Semenišín, Tamara Šejnová, Nikola Šimková, Michal Vosinek, Lucie Zavadilová, and Jan Zibner. Annotated Corpus of Czech Case Law for Reference Recognition Tasks. \textit{Proceedings of TSD 2018}, pp. 239--250.

\bibitem{Hoang}
Duc Tam Hoang, and Ondřej Bojar. TmTriangulate: A Tool for Phrase Table Triangulation. \textit{Prague Bulletion of Mathematical Linguistics}, 2015, no. 104, pp. 75--86.

\bibitem{Hofler}
Stefan Höfler and Michael Piotrowski. Building Corpora for the Philological Study of Swiss Legal Texts. \textit{Journal for Language Technology an Computational Linguistics}, 2011, vol. 26, n. 2, pp. 77--89.

\bibitem{Kriz}
Vincent Kríž, Barbora Hladká, Jan Dědek and Martin Nečaský. Statistical Recognition of References in Czech Court Decisions. \textit{Proceedings of MICAI 2014, Part I}, pp. 51--61.

\bibitem{Perez}
José Marín Peréz and Camino Rea Rizzo. Structure and Design of the British Law Report Corpus (BLRC): A Legal Corpus of Judicial Decisions from the UK. \textit{Journal of English Studies}, 2012, vol. 10, pp. 131--145.

\bibitem{Ro-Pue}
Paula Rodríguez-Puente. Introducing the Corpus of Historical English Law Reports: Structure and compilation techniques. \textit{Revistas de Lenguas para Fines Específicos}, 2011, vol. 17, pp. 99--120.

\bibitem{Smekal}
Katarína Šipulová, Hubert Smekal and Jozef Janovský. Searching for a Reference: Using Automated Text Analysis to Study Judicial Compliance. \textit{Masaryk University Journal of Law and Technology}, 2018, vol. 12, no. 2, pp. 131--160.

\bibitem{Steinberger}
Ralf Steinberger, Mohamed Ebrahim, Alexandros Poulis, Manuel Carrasco-Benitez, Patrick Schlüter, Marek Przybyszewski, and Signe Gilbro. An Overview of the European Union's Highly Multilingual Parallel Corpora. \textit{Language Resources and Evaluation}, 2014, vol. 48, no. 4, pp. 679--707.

\bibitem{Vogel}
Friedemann Vogel, Hanjo Hammann, and Isabelle Gauer. Computer-Assisted Legal Linguistics: Corpus Analysis as a New Tool for Legal Studies. \textit{Law \& Social Inquiry}, 2017, early view.

\bibitem{Walker}
Vern R. Walker. The Need for Annotated Corpora from Legal Documents, and for (Human) Protocols for Creating them: the Attribution Problem. In: \textit{Natural Language Argumentation: Mining, Processing, and Reasoning over Textual Arguments (Dagstuhl Seminar 16161)}, ed. Elena Cabrio, Hirst Graeme, Serena Villata and Adam Wyner. 2016.

\bibitem{Ziemski}
Michał Ziemski, Marcin Junczys-Dowmunt, and Bruno Pouliquen. The United Nations Parallel Corpus. \textit{Proceedings of LREC 2016}, pp. 3530--3534.

\end{thebibliography}

\section*{Annex 1}

\begin{center}
 \begin{table} [H]
  \centering
  \begin{tabular}{l|lll|l}
    \toprule
    \cmidrule(r){1-5}
    Year &ConCo &SupCo &SupAdmCo &All courts\\
    \midrule
    1993 &225 &0 &0 &225   \\
    1994 &626 &0 &0 &626   \\
    1995 &775 &0 &0 &775  \\
    1996 &1038 &0 &0 &1038 \\
    1997 &1368 &22 &0 &1390 \\
    1998 &1628 &41 &0 &1669 \\
    1999 &2080 &44 &0 &2124\\
    2000 &2587 &1949 &0 &4536\\
    2001 &2825 &3380 &0 &6205\\
    2002 &2696 &3991 &0 &6687\\
    2003 &2463 &4877 &600 &7940\\
    2004 &2546 &5167 &748 &8461\\
    2005 &2639 &5605 &4524 &12768\\
    2006 &3246 &5809 &4466 & 13521\\
    2007 &3374 &5715 &4703 &13792\\
    2008 &3420 &6310 &3735 &13465\\
    2009 &3441 &6715 &3500 &13656\\
    2010 &3722 &7031 &2965 &13718\\
    2011 &3712 &7243 &3112 &14067\\
    2012 &4565 &6522 &3142 &14229\\
    2013 &3800 &6176 &3600 &13576\\
    2014 &4395 &6810 &3316 &14521\\
    2015 &3857 &7244 &3613 &14714\\
    2016 &4349 &7740 &3527 &15616\\
    2017 &4353 &8059 &3970 &16382\\
    2018* &3356 &5527 &3139 &12022\\
    \bottomrule
  \end{tabular}
  \label{tab:table}
\end{table}
\end{center}

\section*{Annex 2}

\begin{center}
\begin{tikzpicture}
\begin{axis}[
    ybar,
    enlargelimits=0.15,
    legend style={at={(0.5,-0.15)},
      anchor=north,legend columns=-1},
    ylabel={\# of documents},
    symbolic x coords={1993,1994,1995,1996,1997,1998,1999},
    xtick=data,
    nodes near coords,
    nodes near coords align={vertical},
    ]
\addplot coordinates {(1993,225) (1994,626) (1995,775) (1996,1038) (1997,1368) (1998,1628) (1999,2080)};
\addplot coordinates {(1997,22) (1998,41) (1999,44)};
\legend{ConCo,SupCo}
\end{axis}
\end{tikzpicture}
\end{center}

\begin{center}
\begin{tikzpicture}
\begin{axis}[
    ybar,
    enlargelimits=0.15,
    legend style={at={(0.5,-0.15)},
      anchor=north,legend columns=-1},
    ylabel={\# of documents},
    symbolic x coords={2000,2001,2002,2003,2004,2005,2006,2007,2008,2009},
    xtick=data,
    nodes near coords,
    nodes near coords align={vertical},
    ]
\addplot coordinates {(2000,2587) (2001,2825) (2002,2696) (2003,2463) (2004,2546) (2005,2639) (2006,3246) (2007,3374) (2008,3420) (2009,3441)};
\addplot coordinates {(2000,1949) (2001,3380) (2002,3991) (2003,4877) (2004,5167) (2005,5605) (2006,5809) (2007,5715) (2008,6310) (2009,6715)};
\addplot coordinates {(2003,600) (2004,748) (2005,4524) (2006,4466) (2007,4703) (2008,3735) (2009,3500)};
\legend{ConCo,SupCo,SupAdmCo}
\end{axis}
\end{tikzpicture}
\end{center}

\begin{center}
\begin{tikzpicture}
\begin{axis}[
    ybar,
    enlargelimits=0.15,
    legend style={at={(0.5,-0.15)},
      anchor=north,legend columns=-1},
    ylabel={\# of documents},
    symbolic x coords={2010,2011,2012,2013,2014,2015,2016,2017,2018*},
    xtick=data,
    nodes near coords,
    nodes near coords align={vertical},
    ]
\addplot coordinates {(2010,3722) (2011,3712) (2012,4565) (2013,3800) (2014,4395) (2015,3857) (2016,4349) (2017,4353) (2018*,3356)};
\addplot coordinates {(2010,7031) (2011,7243) (2012,6522) (2013,6176) (2014,6810) (2015,7244) (2016,7740) (2017,8059) (2018*,5527)};
\addplot coordinates {(2010,2965) (2011,3112) (2012,3142) (2013,3600) (2014,3316) (2015,3613) (2016,3527) (2017,3970) (2018*,3139)};
\legend{ConCo,SupCo,SupAdmCo}
\end{axis}
\end{tikzpicture}
\end{center}

\end{document}